\documentclass[conference]{IEEEtran}
\IEEEoverridecommandlockouts
\usepackage{cite}
\usepackage{amsmath,amssymb,amsfonts}
\usepackage{algorithmic}
\usepackage{graphicx}
\usepackage{textcomp}
\usepackage{xcolor}
\usepackage{tabularray} % for tblr
\usepackage{pifont} % for \XSolidBrush series
\usepackage{multirow}
\usepackage{booktabs}
\usepackage{bbding}
\usepackage{soul}     % 用于删除线（\st{} 命令）
\usepackage[hidelinks]{hyperref}
\def\BibTeX{{\rm B\kern-.05em{\sc i\kern-.025em b}\kern-.08em
    T\kern-.1667em\lower.7ex\hbox{E}\kern-.125emX}}
\begin{document}

\title{Moving Light Adaptive Colonoscopy Reconstruction via Illumination-Attenuation-Aware 3D Gaussian Splatting}

\author{
\IEEEauthorblockN{Hao Wang$^{a,\dag}$, Ying Zhou$^{a,\dag}$, Haoyu Zhao$^{b}$, Rui Wang$^{a}$, Qiang Hu$^{a}$, Xing Zhang$^{c}$, Qiang Li$^{a,*}$, Zhiwei Wang$^{a,*}$}
\IEEEauthorblockA{\textit{$^a$Wuhan National Laboratory for Optoelectronics, Huazhong University of Science and Technology} \\
    \textit{$^b$School of Computer Science, Wuhan University}\\
    \textit{$^c$Wuhan United Imaging Healthcare Surgical Technology Co., Ltd}\\
    \{liqiang8, zwwang\}@hust.edu.cn}
}
\maketitle
\begingroup
\renewcommand{\thefootnote}{}
\footnotetext{$^\dagger$ Equal contribution. $^*$ Corresponding authors.}
\endgroup

\begin{abstract}
% 3D Gaussian Splatting (3DGS) enables real-time view synthesis in colonoscopy but assumes static illumination, making it incompatible with the strong photometric variations caused by the moving light source and camera.
% This mismatch leads existing methods to compensate illumination attenuation with structure-violating Gaussians, degrading geometric fidelity.
% Prior work considers only distance-based attenuation and overlooks the physical characteristics of colonscopic lighting. In this paper, we propose ColIAGS, an improved 3DGS framework for colonoscopy. 
% To mimic realistic appearance under varying illumination, we introduce an Improved Appearance Modeling with two types of illumination attenuation factors, enabling Gaussians to adapt to photometric changes while a cosine embedding is used to implicitly generate attenuation solutions. We further design an Improved Geometry Modeling using high-dimensional view embeddings to strengthen geometry details.
% Experimental results on standard benchmarks demonstrate that ColIAGS supports both high-quality novel-view synthesis and accurate geometry reconstruction, outperforming state-of-the-art methods in rendering fidelity and Depth MSE.
3D Gaussian Splatting (3DGS) enables real-time view synthesis in colonoscopy but assumes static illumination, making it incompatible with the strong photometric variations caused by the moving light source and camera.
This mismatch leads existing methods to compensate for illumination attenuation with structure-violating Gaussians, degrading geometric fidelity.
Prior work considers only distance-based attenuation and overlooks the physical characteristics of colonscopic lighting. In this paper, we propose ColIAGS, an improved 3DGS framework for colonoscopy. 
To mimic realistic appearance under varying illumination, we introduce a lighting model with two types of illumination attenuation factors. To satisfy this lighting model’s approximation and effectively integrate it into the 3DGS framework, we design Improved Geometry Modeling to strengthen geometry details and Improved Appearance Modeling to implicitly optimize illumination attenuation solutions.
Experimental results on standard benchmarks demonstrate that ColIAGS supports both high-quality novel-view synthesis and accurate geometry reconstruction, outperforming state-of-the-art methods in rendering fidelity and Depth MSE.
Our code is available at \url{https://github.com/haowang020110/ColIAGS}.
\end{abstract}


\section{Introduction}
\label{sec:intro}
%不改
Colorectal cancer (CRC) remains one of the leading causes of cancer-related deaths globally~\cite{cancer}, and early detection via colonoscopy~\cite{sali,monobox} is crucial for improving survival rates. 
% However, the complex anatomical structure of the colon and rectum, including the plicae circulares, often obstructs the view during colonoscopy, leading to missed detections~\cite{rex2015quality}. 
To address the missed detection challenge, 3D reconstruction of the colon through advanced imaging techniques is essential. A reconstructed 3D model not only enhances visibility through novel view synthesis but also supports critical applications such as surgical planning~\cite{surgicalplan}, training~\cite{lerplane}, and follow-up screenings~\cite{rnnslam}.
%不改

Recent advancements have been made in the field of endoscopic scene recovery. Traditional algorithms, such as simultaneous localization and mapping (SLAM)~\cite{rnnslam} and depth estimation~\cite{zhou2024improved}, often generate 3D point cloud models from feature point extraction. However, these models lack realistic appearance. Neural Radiance Fields~\cite{nerf} (NeRF)-based approaches~\cite{reim,endonerf,lerplane,endoself,dnfplane,ucnerf} achieve photorealistic renderings through ray tracing but are computationally intensive, requiring lengthy training and inference times.
%不改
3D Gaussian Splatting (3DGS)~\cite{3dgs} has emerged as a promising alternative for various application~\cite{zhao2025high}. For endoscopy, it brings significant improvements in both rendering realism and computational efficiency for 3D reconstruction~\cite{free-surgs,pancakes} and 4D video reconstruction~\cite{hfgs,deform3dgs,lgs,endo4dgs}.
%改gaigai
%问题1

However, the vanilla 3DGS assumes static illumination and that observed appearance depends solely on view angle, consequently failing to handle the illumination attenuation inherent in practical colonoscopy scenarios.
To compensate, most existing 3DGS methods generate structure-violating Gaussians, which obscure their behind structure-preserving Gaussians, leading to significant 3D reconstruction errors~\cite{vastgaussian}.
%问题2

Existing approaches~\cite{reim,prendo} have attempted to incorporate light source distance into the appearance modeling of NeRF/GS frameworks, which partially alleviates the aforementioned issues.
However, in endoscopic scenes, illumination attenuation is influenced not only by the distance to the light source, but also by the orientation of both the light source and the camera — factors that these methods fail to account for.

%our method
%p1开头
In this paper, we propose ColIAGS, an improved 3DGS variant that addresses the aforementioned incompatibility issue, enhancing both rendering fidelity and geometric reconstruction accuracy in colonoscopy scenes.
Specifically, we first perform Illumination Factor Extraction on colonoscopy scenes and derive a novel lighting modeling approximation within the 3DGS framework. This model establishes that the observed appearance depends not only on the view angle, but also on two more types of illumination attenuation. 
However, this modeling approximation holds only under the condition that Gaussians sufficiently adhere to the tissue surface and capture structural details. To this end, we introduce an Improved Geometry Modeling with View Embedding for restricted view angles to enhance the Gaussians' geometry attribute prediction, thereby satisfying the approximation criteria. 
To achieve more efficient appearance modeling, we further proposed an Improved Appearance Modeling with Illumination Attenuation, employing an MLP with cosine embedding inputs to implicitly solve the attenuation behavior from high-dimensional feature representations. Consequently, establishes the observed appearance as a function of both camera-to-surface distance and orientation.

Our contributions can be summarized as follows:
% \begin{figure} \includegraphics[width=1.0\linewidth]{Fig/fig1.png} 
% \centering 
% \caption{(a) The tissue appearance observed from the same view direction but with different light source positions. (b) The structure-violating vaporous Gaussian blobs (in black) generated along the motion trajectories to simulate dynamic lighting variations.} \label{fig1} 
% \end{figure}

\begin{enumerate} 
\item We propose ColIAGS, an improved 3DGS variant that improves both rendering fidelity and 3D reconstruction accuracy significantly.
\item We introduce a lighting model tailored for the colonoscopic scenario and intergrate it into 3DGS with two novel improved modeling schemes to satisfy model's approximation.
\item Experiments on two public benchmarks demonstrate the superiority of ColIAGS over state-of-the-art methods.
% (1) it maintains comparable rendering fidelity while significantly improving geometric accuracy relative to realistic rendering techniques, and 
% (2) it achieves a 2.04 dB improvement in PSNR for novel view synthesis and reduces Depth MSE by 78\% compared to geometry-preserving approaches. 
\end{enumerate}

\begin{figure}[!t]
    \includegraphics[width=1.0\linewidth]{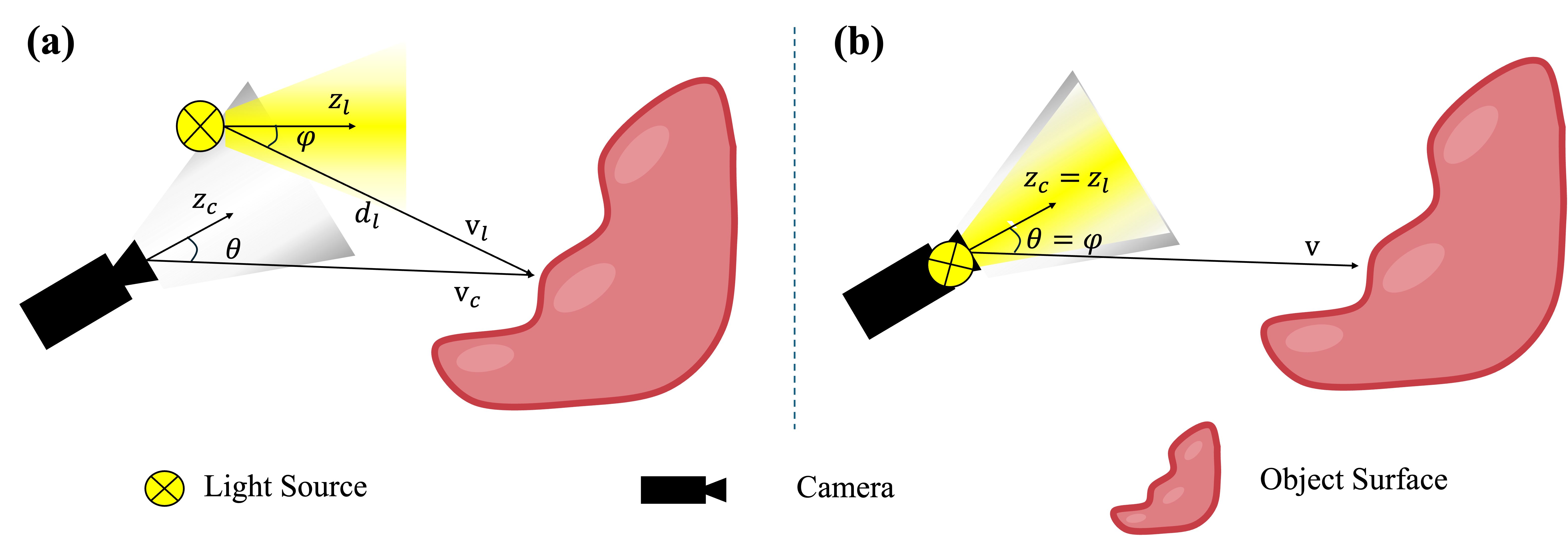}
\caption{(a) illustrates the illumination attenuation in a conventional point light model, where the illumination depends on both the light source’s orientation and its distance to the object surface, while camera vignetting effects are influenced by the view direction.
(b) presents a simplified lighting model, where the light source is approximated to be co-located with the camera, and its direction is assumed to align with the camera's orientation. } \label{lightmodel}
\end{figure}

\begin{figure*}[!t]
\includegraphics[width=1.0\linewidth]{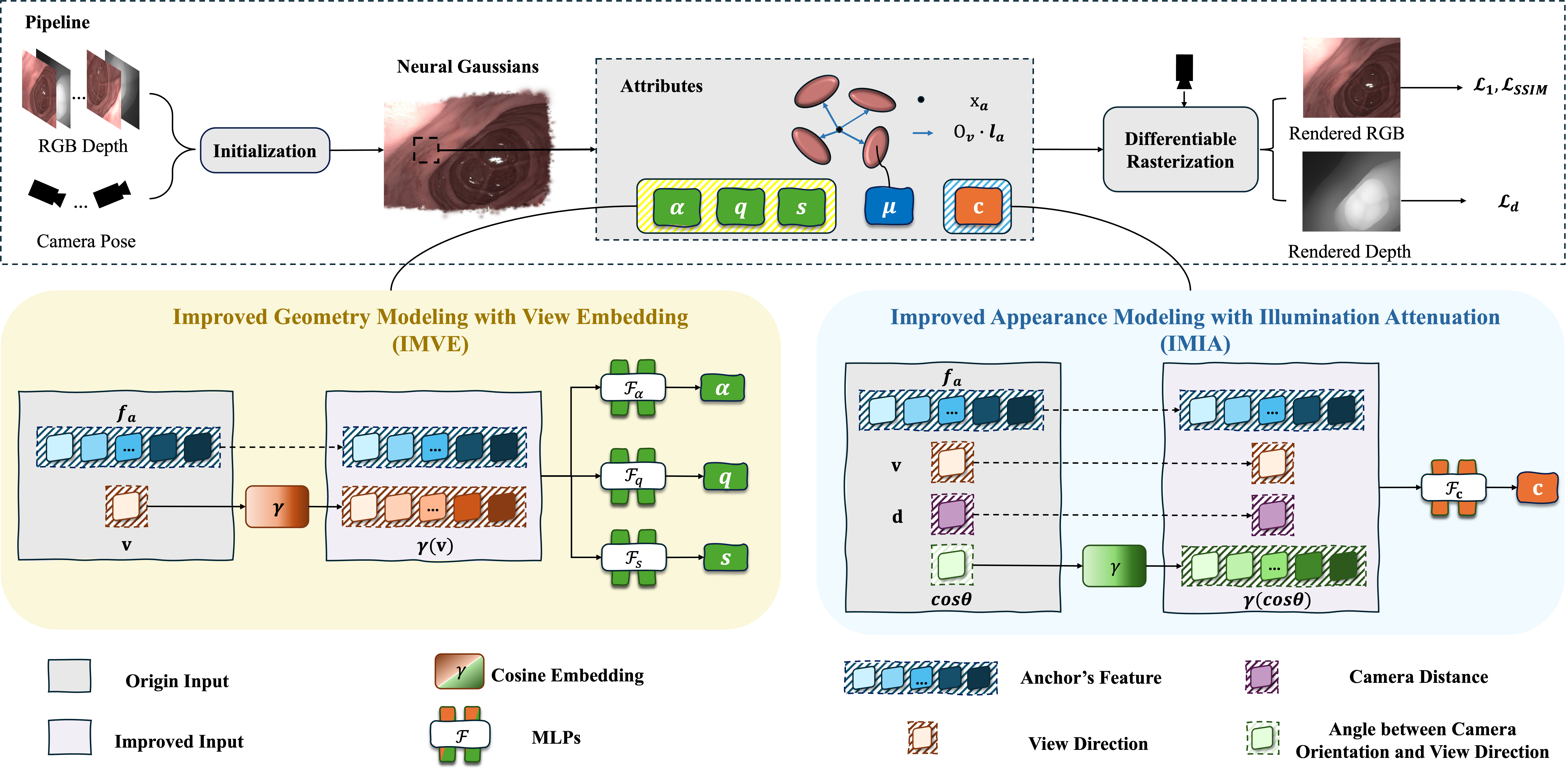}
\caption{Overview of our proposed ColIAGS framework. The pipeline of ColIAGS contains two proposed components, i.e., the Improved Geometry Modeling with View Embedding(the yellow box) to enhance the geometry precision, ensuring the approximation conditions to incorporate illumination factors, and the Improved Appearance Modeling with Illumination Attenuation(the blue box) to model the two types of illumination attenuation. } \label{fig1:method}
\end{figure*}

\section{Method}
\subsection{Preliminaries of 3DGS and Neural Gaussians}
\label{2.1}

\subsubsection{3DGS}
Images are 2D observations of a 3D real-world scene. As illustrated in the top box of Fig.~\ref{fig1:method}, 3D Gaussian Splatting (3DGS) represents the underlying scene using a set of anisotropic 3D Gaussians $\mathbf{\Theta} = \{ \mu, s,q, \alpha, \mathbf{c} \}$, where $\mu$ is the mean position, $s, q$ are the scaling matrix and rotation quaternion components of 3D covariance matrix,
$\alpha$ is the opacity, and $\mathbf{c}$ is the Gaussian's color defined by the spherical harmonics to model view-variant color.

3DGS efficiently renders the scene using tile-based rasterization. First, the 3D Gaussians are projected onto the 2D image plane~\cite{ewa}. Then, the 2D Gaussians are sorted, and $\alpha$-blending is applied: 
\begin{equation} 
\label{eq:aplha blend} 
C(x') = \sum_{i\in{N}}\mathbf{c}_i\sigma_i\prod_{j=1}^{i-1}(1-\sigma_j) 
\end{equation} where $x'$ is the queried pixel position, $N$ denotes the number of sorted 2D Gaussians associated with the queried pixel, and $\sigma$ is the 2D Gaussian opacity defined by its 3D counterpart $\alpha$. By leveraging a differentiable rasterizer, $\mathbf{\Theta}$ is learnable and optimized end-to-end via view reconstruction training.

In addition, the gaussian-surface distance can be approximated by rendering their depths similarly to Eq.~(\ref{eq:aplha blend}), as follows:
\begin{equation} 
D(x') = \sum_{i\in{N}} z_i \sigma_i \prod_{j=1}^{i-1}(1-\sigma_j), \quad z_i = \mathbf{R}^{-1}(\mu_i-\mathbf{x}_c)_z
\end{equation}
% where the rendered depth $D(x')$, which reflects the weighted summed distance of the sorted Gaussians' positions along the view direction. $D(x')$ is rendered similarly to Eq.~(\ref{eq:aplha blend}), as follows: 
where $z_i$ is the $i$-th Gaussian's depth~\cite{depthgs} in the camera coordinate system. $\mathbf{R}$ and $\mathbf{x}_c$ are the camera rotation and position, and $( \cdot)_z$ means fetching the coordinate's $z$-value.

% \begin{figure*}
%     \includegraphics[width=1.0\linewidth]{Fig/fig2_v3.png}
% \caption{(a) illustrates the lighting attenuation in a conventional point light model, where the illumination depends on both the light source’s orientation and its distance to the object surface, while camera vignetting effects are influenced by the viewing direction.
% (b) presents a simplified lighting model, where the light source is approximated to be co-located with the camera, and its direction is assumed to align with the camera's orientation. } \label{fig2}
% \end{figure*}

\subsubsection{Neural Gaussians}
The anchor-based neural Gaussian technique was originally proposed by~\cite{scaffold} for efficient on-the-fly rendering and redundancy reduction via feature-enriched anchors. 
% We adapt this technique for colonoscopy scenes to address tissue color variations caused by the moving light source.We first obtain a sparse colon point cloud by reprojecting the depth map into world space via camera pose.
The scene is then hierarchically voxelized, with each voxel center assigned an anchor containing a context feature $f_a$, a scaling factor $l_a \in \mathbb{R}^{3}$, and $k$ learnable offsets $\mathbf{O}_v \in \mathbb{R}^{k\times 3}$. 
Anchors within the viewing frustum dynamically generate $k$ Gaussians, whose positions are calculated by:
\begin{equation}
    \mu = \mathbf{x}_a + \mathbf{O}_v \cdot l_a
\end{equation}
while other attributes in $\mathbf{\Theta}$ are predicted by individual MLPs with $f_a$, view distance $\mathbf{d}=\Vert \mathbf{x}_{a}-\mathbf{x}_{c} \Vert$, and direction $\mathbf{v}=(\mathbf{x}_{a}-\mathbf{x}_{c})/\mathbf{d}$ pointing from the camera position $\mathbf{x}_{c}$ to the anchor's $\mathbf{x}_{a}$ as input.

% Therefore, the 3D Gaussians' geometry attributes, such as shape and opacity (denoted by $S$, $R$, and $\alpha$), are no longer trained individually but are managed collectively through anchor features, regularizing their placement to anchored regions. This \textbf{Anchor-based Regularization} prevents excessive clustering on the colon surface and encourages movement toward less-observed areas for better surface recovery. 
Both GS and Neural Gaussians are designed for the natural scene scenario. 
However, natural scenes typically have wide fields of view and static lighting conditions, which simplify the reconstruction process. In contrast, the colonoscopy is characterized by light sources that are physically attached to the endoscope itself. 
The vanilla application of these methods on the colonoscopic scene leads to geometrically implausible reconstructions and inaccurate modeling of lighting.
To this end, we establish a specialized model tailored to the complex illumination of colonoscopy and implement it in an efficient manner.

\subsection{Lighting Modeling and Illumination Attenuation Factor Extraction}
In contrast to natural scenarios, where view angle is the dominant factor for appearance, endoscopic scenes exhibit appearance that is further determined by two types of illumination attenuation.
The first arises from the movement of the light source, where tissues closer to the light source appear brighter than those farther away. 
The second results from the physical characteristics of both the light source and the camera imaging system, including light spread function and vignetting effects.%figab

As shown in Fig.~\ref{lightmodel}(a), the image color is not only negatively correlated with the light source distance $\mathbf{d}_l$, but also related to the light spread function and the camera vignetting effects, which can be specified by $\mu(x)$ and $V(x)$:
\begin{equation}
\begin{aligned}
    &\mu(x)=\cos^{k_1}(\psi), \psi=<\mathbf{v}_l, \mathbf{z}_l>\\
    &V(x)=\cos^{k_2}(\theta), \theta=<\mathbf{v}_c, \mathbf{z}_c>\\
    & C \propto \mathbf{d}_l^{-1},\mu(x),V(x)
\end{aligned}
\end{equation}
where $\psi$/$\theta$ are the angle between light/camera view direction $\mathbf{v}_l$/$\mathbf{v}_c$ and light/camera orientation $\mathbf{z}_l/\mathbf{z}_c$, $k1,k2$ are the exponents related to the physical characteristics, and $\mathbf{d}_l$ is the distance between the light source and the surface of the object.
% In other words, the closer the light source is, the smaller the $\theta$ and $\psi$ angles are, and the brighter the pixel appears.

Inspired by~\cite{photometric}, the lighting model in colonoscopy can be approximated by assuming the light source is attached to the camera, as illustrated in Fig.~\ref{lightmodel}(b).

To extract illumination attenuation factors in our Neural Gaussians framework, we approximate object's surface with Gaussians, using $\mathbf{v}$/$\mathbf{d}$ as the direction/distance from camera/light to object surface.
% These approximation allow us to model light source movement-induced variation as a function of camera distance $\mathbf{d}$, i.e. $\mathbf{d}_l=\mathbf{d}$. 
Furthermore, attenuation effects caused by both the light spread function and the camera vignetting effects can be modeled as a unified relationship $\mu'(x)$ with Gaussians' color $\mathbf{c}$, i.e., $\psi=\theta$:
\begin{equation}
    \mathbf{c} \propto \mu'(x)=\mu(x)V(x)=\cos^{k}{\theta}
\end{equation} 
where $\theta$ is the angle between view direction $\mathbf{v}$ between view direction and light/camera's orientation and $k$ is an undetermined exponent.

However, the imperfect geometry attributes of Gaussians leave gap between the assumption and reality, i.e., Gaussians cannot properly adhere to object surfaces or is not fine enough to present structure details.
Additionally, since the exponential parameter k varies between different physical characters, directly estimate it will be hindered by camera's auto gain as shown in~\cite{photometric}.

\begin{table*}[!t]
\centering
\caption{Quantitative comparison results on C3VD and RotateColon. The best results are marked in bold and the second-best underlined.}

\begin{tabular}{lcccc|ccc}
\toprule
\multirow{2}{*}{Methods} & \multicolumn{4}{c|}{C3VD} & \multicolumn{3}{c}{RotateColon} \\
\cmidrule(r){2-5} \cmidrule(r){6-8}
& Depth MSE $\downarrow$ & PSNR $\uparrow$ & SSIM $\uparrow$ & LPIPS $\downarrow$   
& PSNR $\uparrow$ & SSIM $\uparrow$ & LPIPS $\downarrow$ \\
\midrule
REIM NeRF~\cite{reim} 
& 1.480 & 33.96 & 0.86 & 0.32
&  10.96 &  0.62 & 0.57\\
2DGS~\cite{2dgs}
& 4.839 & 32.93 & 0.88 & 0.24
&  19.58 & 0.87 & 0.29 \\
Gaussian Pancakes~\cite{pancakes} 
& 1.222 & 32.93 & 0.88 & 0.24
& 20.10 & 0.88 & 0.27 \\
3DGS~\cite{3dgs}
& 0.355 & 34.10 & \underline{0.89} & 0.22
& 20.53 & 0.87 & 0.26 \\
Scaffold-GS~\cite{scaffold}
& \underline{0.195} & \underline{34.26} & \underline{0.89} & \underline{0.21}
& \underline{22.58} & \underline{0.89} & \underline{0.22} \\
PR-Endo~\cite{prendo}
& 0.674 & 34.15 &\underline{0.89} & 0.23
& 21.90 & 0.87 & 0.28 \\
Ours 
& \textbf{0.042} & \textbf{36.30} & \textbf{0.91} & \textbf{0.15} 
& \textbf{23.29} & \textbf{0.90} & \textbf{0.21} \\
\bottomrule
\end{tabular}

\label{tab1}
\end{table*}

\begin{table}[!t]
\centering
\caption{Quantitative comparison results on C3VD with EndoGSLAM Initialization. The best results are marked in bold and the second-best underlined.}
\resizebox{\linewidth}{!}
{
\begin{tabular}{lcccc}
\toprule
Method& Depth MSE $\downarrow$ & PSNR $\uparrow$ & SSIM $\uparrow$ & LPIPS $\downarrow$ \\
\midrule
REIM NeRF~\cite{reim} 
& 0.495 & \underline{34.34} & 0.88 & 0.34\\
2DGS~\cite{2dgs}
& 0.682 & 32.05 & 0.88 & 0.34\\
Gaussian Pancakes~\cite{pancakes} 
& 1.630 & 32.28 & 0.88 & 0.32  \\
3DGS~\cite{3dgs}
& 0.333 & 33.56 & \underline{0.89} & \underline{0.29}\\
Scaffold-GS~\cite{scaffold}
& \underline{0.215} & 33.61 & \underline{0.89} & \underline{0.29}\\
PR-Endo~\cite{prendo}
& 0.351 & 34.00 & \underline{0.89} & \underline{0.29}\\
Ours  
& \textbf{0.122} & \textbf{35.58} & \textbf{0.91} & \textbf{0.23}\\
\bottomrule
\end{tabular}
  }
\label{tab2}
\end{table}

\subsection{Improved Modeling with Cosine Embedding}
To accommodate the aforementioned lighting model, we primarily improve the geometry modeling in Neural Gaussians to enable the desired approximation.
Firstly, we leverage depth loss to constrain the position of Gaussians close to object surface:
\begin{equation} 
\mathcal{L}_{d} = \sum_{x'} |D(x') - \hat{D}(x')| 
\label{eq:depth}
\end{equation}

% In colonoscopic scene reconstruction, the limited field of view significantly contributes to the generation of structure-violating Gaussians. Specifically, the restricted viewpoint makes it difficult for the Gaussians to model high-frequency details. 
% Instead, they resort to using large-scale Gaussians to represent low-frequency information, such as blurred or over-smoothed regions.
% However, the limited field of view significantly contributes to the generation of coarse Gaussians i.e. using large-scale Gaussians to represent low-frequency information, such as blurred or over-smoothed regions, hindering the follow-up appearance modeling.
However, optimizing fine structures within a limited view range (which is common in colonoscopy) often leads to losses of structure details~\cite{nerf}, i.e., using large-scale Gaussians to represent low-frequency information, such as blurred or over-smoothed regions.

To enhance the high-frequency representational capability of Neural Gaussians' geometry attributes, we introduce a cosine-based embedding function $\gamma$ into geometry modeling,
which projects the original input $\mathbf{v}$ into a high-dimensional space using a set of cosusoidal basis functions. 
The \textbf{Improved Geometry Modeling with View Embedding (IMVE)} shown in Fig.~\ref{fig1:method}(bottom left) can be expressed as follows:
\begin{equation}
\begin{aligned}
&\alpha = \mathcal{F}_{\alpha}(f_a,\gamma(\mathbf{v})), \\
&s = \mathcal{F}_{s}(f_a,\gamma(\mathbf{v})),\\
&r = \mathcal{F}_{r}(f_a,\gamma(\mathbf{v})). \\
\end{aligned}
\end{equation}
% It is also worth noting that, unlike Scaffold-GS~\cite{scaffold}, we decouple the relationship between camera distance $\mathbf{d}$ and the geometric attributes of Neural Gaussians. 
% This design prevents the model from indirectly fitting tissue's appearance by adjusting geometric attributes, which could otherwise lead to degradation of the geometry.

Given the improved geometry modeling, we can directly model lighting based on the approximation derived from Equation (5). 
% The remaining challenge lies in estimating the unknown exponent k.
% Instead of explicitly estimating the unknown exponent k, we also project employ the ability of the MLP that implicitly learn to model the attenuation behavior from the high-dimensional representation.
Instead of explicitly estimating the unknown exponent k, we employ the ability of the MLP that implicitly learn to mimic the attenuation behavior from the high-dimensional representation.
While similarly based on the cosine embedding, the \textbf{Improved Appearance Modeling with Illumination Attenuation (IMIA)} shown in Fig.~\ref{fig1:method}(bottom right) distinctively incorporates $\theta$ into the input of $\mathcal{F}_c$:
\begin{equation}
\mathbf{c}=\mathcal{F}_c(f_a, \mathbf{v}, \mathbf{d}, \gamma(\cos{\theta}))\\
\end{equation}
The color $\mathbf{c}$ of each Gaussian is now influenced by both camera (light) distance and orientation, accurately modeling the two types of illumination attenuation in colonoscopy. 

\subsection{Loss Function}
\label{2.3}
To train our ColIAGS, we use the image reconstruction losses~\cite{3dgs}, i.e., $\mathcal{L}_{1}$ and $\mathcal{L}_{D-SSIM}$, as well as the above depth constraint $\mathcal{L}_{d}$. 
An extra regularization $\mathcal{L}_{scale}$~\cite{scaleloss} is applied to prevent the overlapping and large volume of Gaussians. 

% The attributes of 3DGS can be divided into two parts: geometry($\{ \mu, \Sigma, \alpha\}$ and appearance $SH$.
% First,
% and perform depth constraint calculations
% We first leveapproximate the Gaussian-surface distance by rendering their depths.

The total constraints are calculated as follows:
\begin{equation}
\mathcal{L} = (1 - \lambda_1)\mathcal{L}_{1} + \lambda_1\mathcal{L}_{D-SSIM} + \lambda_2\mathcal{L}_{depth} + \lambda_3\mathcal{L}_{scale}
\end{equation}

\section{Experiment Settings}

\subsection{Implementation Details}
We train ColIAGS using Pytorch framework on a single NVIDIA GeForce RTX 4090. 
Following the previous settings~\cite{3dgs,depth_hyper,scaffold}, we set $\lambda_1$, $\lambda_2$, $\lambda_3$ as 0.2, (0.2-0.01 with exponential decay), 0.01, respectively.
The cosine embedding strategies in IMVE and IMIA are configured with dimensions of 10 and 5, respectively. %($D_\mathbf{v}=10$ $D_\theta=5$).
More details can be found in our released code.
% For other setting about neural gaussians, we follow the default settings in~\cite{scaffold}.

\subsection{Dataset and Evaluation Metrics}
We follow the protocol of existing study \cite{reim,pancakes,prendo}, conduct evaluation on C3VD \cite{c3vd}, C3VD with EndoGSLAM Initialization~\cite{endogslam}, and RotateColon~\cite{prendo}.

\subsubsection{C3VD}
consists of 22 colonoscopy videos with the resolution of $1350 \times 1080$ captured from various colon phantom models, equipped with registered depth maps and corresponding camera poses. These models exhibit the mentioned illumination variations, thus ensuring a comprehensive setup for evaluating the effectiveness of our method. 
In practice, we undistorted and resized the image to $338 \times 270$~\cite{pancakes}. For each scene’s video, we split the frame data into training and testing sets using a 7: 1 ratio.

\subsubsection{RotateColon}
is developed specifically to evaluate novel view synthesis under extended rotations proposed by previous work~\cite{prendo}, including intense rotations not observed during the training sequence.
The per-frame's depth and camera are recorded with their in-house simulator.
We follow the original train/split of prior work~\cite{prendo} for fair comparision.

% Note that, the original settings on the two datasets mentioned above utilize ground-truth camera poses for initialization.
% To further evaluate the robustness and practical applicability of our method, we follow previous work~\cite{prendo,pancakes} by using SLAM outputs as the initialization for GS model.
% To avoid the potential impact of minor discrepancies in SLAM results, we utilize the output released by previous work~\cite{prendo} for consistency.
% It contains 10 sequences selected from C3VD as~\cite{endogslam}, and the camera poses and depth maps used are generated by the EndoGSLAM~\cite{endogslam} framework. This setting validates algorithm feasibility under suboptimal initialization by deliberately adopting SLAM-based initialization, replicating the challenging conditions in Gaussian Pancakes~\cite{pancakes}.
While the original benchmarks utilize ground-truth poses, we follow~\cite{prendo,pancakes} to further evaluate robustness using SLAM-based initialization. 
Specifically, we adopt the evaluation protocol from~\cite{prendo}, using 10 C3VD sequences with poses and depth maps generated by EndoGSLAM~\cite{endogslam}. 
This replicates the challenging conditions of suboptimal initialization as explored in Gaussian Pancakes~\cite{pancakes}.

% \subsubsection{Metrics}
% To be consistent with existing works~\cite{reim,pancakes,prendo}, we thoroughly evaluate the model performance using comprehensive metrics, including peak signal-to-noise ratio (PSNR), structural similarity index (SSIM) and learned perceptual
% image patch similarity (LPIPS) to measure the rendering quality, and depth mean square error (Depth MSE) for geometry accuracy.
% It's notable that as RotateColon did not provide depth ground-truth for test views, we only evaluate Depth MSE on C3VD and C3VD with EndoGSLAM Initialization.
We adopt standard metrics from~\cite{reim,pancakes,prendo}: PSNR, SSIM, and LPIPS for rendering quality, alongside Depth MSE for geometry.
Due to the absence of depth ground-truth in RotateColon, we only evaluate Depth MSE for C3VD.

% Moreover REIM-NeRF's source code has bug in Depth MSE computation, which is reported recently. Therefore, it's results is different from origin paper.

\begin{figure*}[!t]
\centering
\includegraphics[width=\linewidth]{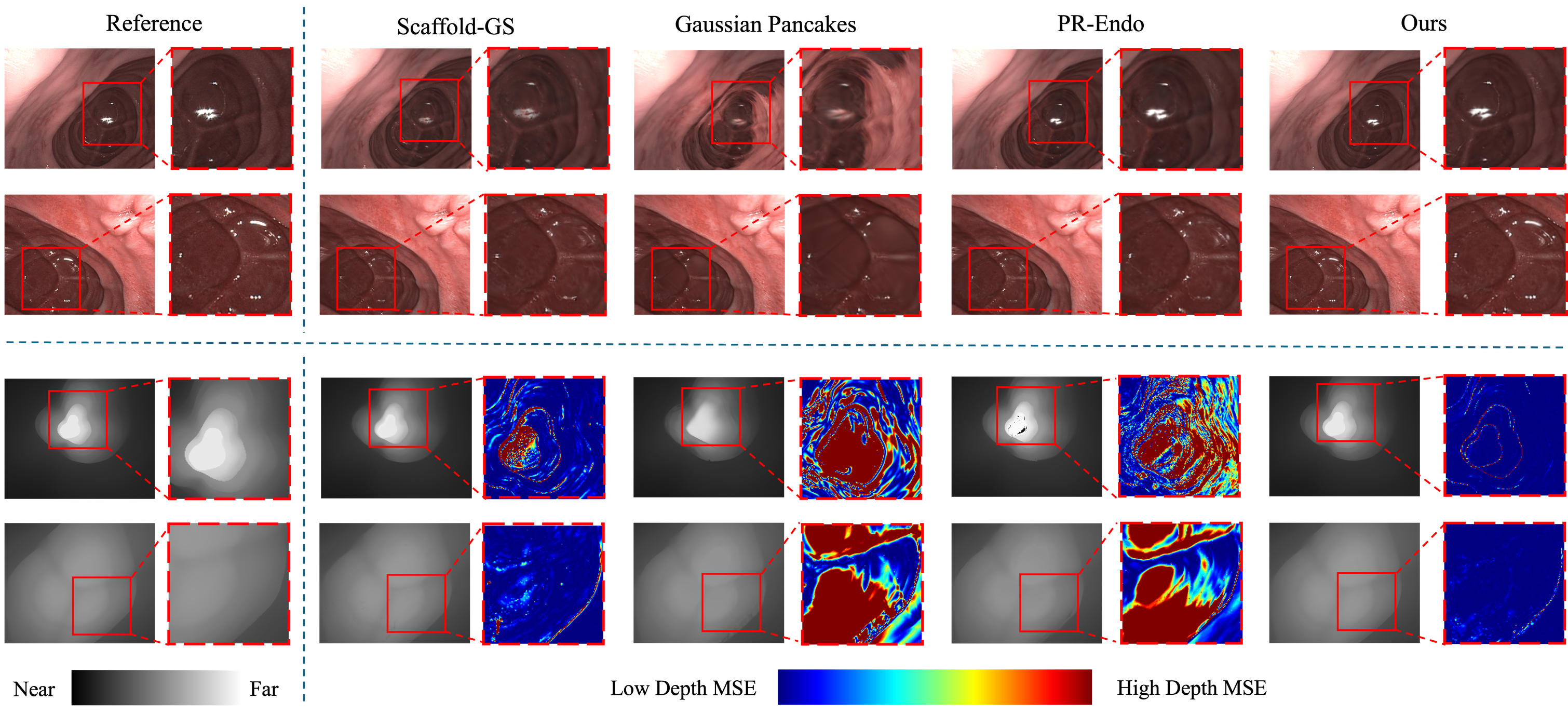}
\caption{Qualitative comparison on C3VD, against Scaffold-GS~\cite{scaffold}, Gaussians Pancakes~\cite{pancakes} and PR-Endo~\cite{prendo}.The top two rows display the rendered RGB images, while the bottom two rows show the depth mse error map.} \label{fig3}
\end{figure*}

\section{Results and Discussion}
We compare ColIAGS with 6 state-of-the-art methods, 
i.e., 3DGS~\cite{3dgs}, 2DGS~\cite{2dgs}, Scaffold-GS~\cite{scaffold}, REIM NeRF~\cite{reim}, Gaussian Pancakes~\cite{pancakes} and PR-Endo~\cite{prendo}.
We evaluate their performance and make comparisons using their released codes. Note that, to ensure fairness in comparisons, we also incorporate depth loss $\mathcal{L}_d$ in Eq.~\ref{eq:depth} to 3DGS and Scaffold-GS, which is a necessary condition for preserving fundamental geometry.

\subsection{Comparison with State-of-the-arts}
Table~\ref{tab1} presents the comparison results between ColIAGS and other state-of-the-art methods. 
We can observe that our method consistently outperforms other methods on both C3VD and RotateColon benchmarks, demonstrating superior rendering quality and geometric accuracy.

Among these SOTA methods, Scaffold-GS and PR-Endo achieve the second and third performance. While both are specifically designed for complex scenes, with PR-Endo additionally incorporating endoscopic-specific considerations, e.g., constrained camera trajectories and view-dependent lighting, their suboptimal performance ultimately derives from insufficient geometric optimization. In contrast, ColIAGS surpasses them with a reduction of 0.155 $mm^2$ in Depth MSE and an improvement of 2.04 dB in PSNR for rendering quality.

To validate robustness under realistic clinical conditions, i.e., the cases without the ground truth, we specifically evaluate the scenario where depth maps and camera poses generated by EndoGSLAM are used for initialization. As shown in Table~\ref{tab2}, our method maintains superior performance in all metrics, which indicates strong potential for clinical deployment.

% Fig~\ref{fig3} visualizes the performance comparison in novel view synthesis and geometric reconstruction.  
% As illustrated in the top two rows, ColIAGS achieves realistic rendering quality by accurately modeling illumination attenuation and preserving sharper high-frequency details, e.g., irregular highlight patterns and fold discontinuities.
% The visualization comparison in the bottom two rows reveals that our method generates more smoothed depth maps containing reduced noise artifacts and higher fidelity to the ground truth depth, confirming the necessity of modeling illumination attenuation and incorporating cosine embeddings.
Fig.~\ref{fig3} illustrates our superiority in both rendering and geometry. 
The top rows show that ColIAGS captures realistic illumination and sharp details. 
Meanwhile, the bottom rows reveal that our method yields smoother, low-noise depth maps close to ground truth, confirming the necessity of our proposed illumination and embedding modules.
% As illustrated in the top two rows, ColIAGS achieves superior rendering quality compared to Gaussian Pancakes, which, due to its enforced depth constraints, exhibits significantly different illumination and visible artifacts.
% The bottom two rows reveal that our method achieve better geometry accuracy compared to others.

\begin{table}[!t]
\centering
\caption{Ablation Study on C3VD~\cite{c3vd}. \textbf{$\mathcal{L}_{depth}$}, \textbf{IMVE} and \textbf{IMIA} refer to depth loss, Improved Geometry Modeling with View Embedding and Improved Appearance Modeling with Illumination Attenuation, respectively.}\label{tab3}
\resizebox{\linewidth}{!}
{
\begin{tblr}{
  cells = {c},
  hline{1,7} = {-}{0.08em},
  hline{2} = {-}{},
}
$\mathcal{L}_{d}$ & IMVE & IMIA & Depth MSE $\downarrow$& PSNR $\uparrow$& SSIM $\uparrow$& LPIPS $\downarrow$\\
\XSolidBrush&\XSolidBrush&\XSolidBrush& 147.644&34.92&0.90  &0.20\\
\Checkmark&\XSolidBrush&\XSolidBrush& 0.195 & 34.26 & 0.89 & 0.21\\
\Checkmark&\XSolidBrush&\Checkmark&0.135 & 35.18	&0.90	&0.18\\
\Checkmark&\Checkmark&\XSolidBrush&0.064 &36.01	&0.90	&0.17 \\
\Checkmark&\Checkmark&\Checkmark&0.042 &36.30&0.91&0.15 \\    
\end{tblr}%
}
\end{table}

% \begin{table}
% \centering
% \caption{Ablation Study on C3VD~\cite{c3vd} about embedding dimensions on $\mathbf{v}$ and $\theta$, respectively.The best results are marked in bold.
% }\label{tab4}
% \begin{tblr}{
%   cells = {c},
%   hline{1,8} = {-}{0.08em},
%   hline{2} = {-}{},
%   hline{5} = {-}{},
% }
% $D_\mathbf{v}$ &$D_\theta$ &Depth MSE$\downarrow$& PSNR $\uparrow$& SSIM $\uparrow$& LPIPS $\downarrow$\\
% 10&0  &0.045 &36.08&0.90&0.16 \\
% 10&5  &\textbf{0.042} &\textbf{36.30}&\textbf{0.91}&\textbf{0.15} \\
% 10&10 &0.051 &36.29&0.91&0.16\\
% 5 &5  &0.057 &36.07&0.91&0.17  \\
% 10&5  &\textbf{0.042} &\textbf{36.30}&\textbf{0.91}&\textbf{0.15} \\
% 15&5  &0.087 &36.06&0.90&0.16 \\
% \end{tblr}%
% \end{table}

\subsection{Ablation Study}
% To further verify the effectiveness of the two proposed techniques tailored for colonoscopic 3DGS, i.e., Improved Geometry Modeling with View Embedding (IMVE) and Improved Appearance Modeling with Illumination Attenuation (IMIA), we conduct ablation study on C3VD. 
% Also, we investigate the impact of the cosine embedding dimensions (dim) in both modules.

% \subsubsection{Effectiveness of Two Components} 
We develop four variants by enabling/disabling each component. Specifically, when the improved modeling is disabled, we use the original input mode in the vanilla Scaffold-GS.
In addition, we develop a variant without the depth constraint $\mathcal{L}_{depth}$, aiming to indicate the necessity of applying it to guarantee basic geometry. 

Table~\ref{tab3} shows the quantitative comparison between the complete ColIAGS and other variants. Based on these results, three key observations can be made as follows:

(1) When comparing the $1^{th}$ to $2^{th}$ rows, the absence of depth supervision leads to a substantial decrease in geometric accuracy.
This justifies our experimental setting to incorporate the depth loss into other variants for comparison, ensuring a fair evaluation of our proposed improvements.
(2) As can be seen from the $2^{th}$ to the $4^{th}$ rows of Table~\ref{tab3}, using either IMVE or IMIA can bring significant improvements in both rendering quality (in terms of PSNR, $p < 0.016$) while IMVE also improve geometry precision (in terms of Depth MSE, $p<0.03$).
(3) A comparison between the $3^{th}$ and $4^{th}$ rows reveals that although incorporating IMIA alone (without IMVE) leads to a certain improvement in rendering quality, the suboptimal geometry accuracy induces notable approximation errors, which ultimately undermine both the rendering results and the overall geometric fidelity. 
(4) The bottom two rows of Table~\ref{tab3} indicate that IMVE and IMIA are not mutually excluded, bringing a further reduction in Depth MSE by at least 78\%. 
Therefore, ColIAGS is able to acquire a model that effectively combines novel view synthesis and geometric reconstruction.

\section{Conclusion}
We presented ColIAGS, an enhanced 3DGS framework tailored for high-fidelity colonoscopic reconstruction. 
ColIAGS addresses the limitations of dynamic illumination and geometric inaccuracy through two key modules:
Improved Geometry Modeling with View Embedding (IMVE), which refines geometric details via high-dimensional view embedding, 
and Improved Appearance Modeling with Illumination Attenuation (IMIA), which explicitly models light attenuation based on distance and orientation. 
Extensive evaluations on the benchmarks demonstrate that 
% ColIAGS outperforms six state-of-the-art methods, achieving a 2.04 dB gain in PSNR and a 78\% reduction in Depth MSE. 
ColIAGS outperforms six state-of-the-art methods. 
Future work will extend this methodology to pose-free paradigms.
% In conclusion, we present ColIAGS, an enhanced 3D Gaussian Splatting (3DGS) framework that overcomes the limitations of vanilla 3DGS in colonoscopic reconstruction by effectively modeling dynamic illumination variations while preserving geometric accuracy. Unlike existing methods, which only consider the light source distance, our method introduces illumination-aware appearance modeling with two attenuation factors and enhances geometry precision through high-dimensional view embedding. 
% Specifically, ColIAGS consists of two key modules, i.e., Improved Geometry Modeling with View Embedding (IMVE) and Improved Appearance Modeling with Illumination Attenuation (IMIA).
% IMVE enhances geometric representation by introducing high-frequency details, thereby improving the accuracy of appearance modeling.
% IMIA incorporates both the camera (or light source) distance and orientation to accurately model the two types of illumination attenuation observed in colonoscopy while implicitly optimizing the illumination attenuation solutions through an MLP.
% The comprehensive comparisons with six state-of-the-art methods on two public benchmarks, namely C3VD and RotateColon, demonstrate that ColIAGS achieves superior performance in both novel view synthesis and geometry precision, significantly improving rendering fidelity with a PSNR gain of 2.04 dB while reducing Depth MSE by 78\%, making ColIAGS a promising technique for high-fidelity colonoscopy applications. 
% In the future, we will explore the field of pose-free paradigm while incorporating the improved modeling in this paper.

\section{Acknowledgment}
This work was supported in part by National Natural Science Foundation of China (Grant No.62576145), National Key R\&D Program of China (Grant No.2023YFC2414900) and research grants 
from Wuhan United Imaging Healthcare Surgical Technology Co., Ltd.
\bibliographystyle{IEEEtran}
\bibliography{IEEEabrv,icme2026references}

\end{document}